\documentclass[sn-mathphys,Numbered]{sn-jnl}%
\usepackage{graphicx}%
\usepackage{multirow}%
\usepackage{amsmath,amssymb,amsfonts}%
\usepackage{amsthm}%
\usepackage{mathrsfs}%
\usepackage[title]{appendix}%
\usepackage{xcolor}%
\usepackage{textcomp}%
\usepackage{manyfoot}%
\usepackage{booktabs}%
\usepackage{algorithm}%
\usepackage{algorithmicx}%
\usepackage{algpseudocode}%
\usepackage{listings} 
\usepackage{blindtext}
\usepackage{geometry}
\usepackage[normalem]{ulem}

\usepackage{enumitem}
\setlist[description]{leftmargin=4\parindent,labelindent=4\parindent}

\theoremstyle{thmstyleone}
 
\begin{document}

\title{\bf Human-AI Coevolution}

\author*[1]{\fnm{Dino} \sur{Pedreschi}}\email{dino.pedreschi@unipi.it}

\author*[2,3]{\fnm{Luca} \sur{Pappalardo}}\email{luca.pappalardo@isti.cnr.it}

\author*[4]{\fnm{Emanuele} \sur{Ferragina}}\email{emanuele.ferragina@sciencespo.fr}

\author[5]{\fnm{Ricardo} \sur{Baeza-Yates}}

\author[5]{\fnm{Albert-László} \sur{Barabási}}

\author[6]{\fnm{Frank} \sur{Dignum}}

\author[6]{\fnm{Virginia} \sur{Dignum}}

\author[5]{\fnm{Tina} \sur{Eliassi-Rad}}

\author[3]{\fnm{Fosca} \sur{Giannotti}}

\author[7]{\fnm{János} \sur{Kertész}}

\author[8]{\fnm{Alistair} \sur{Knott}}

\author[9]{\fnm{Yannis} \sur{Ioannidis}}

\author[10]{\fnm{Paul} \sur{Lukowicz}}

\author[2]{\fnm{Andrea} \sur{Passarella}}

\author[11]{\fnm{Alex Sandy} \sur{Pentland}}

\author[12]{\fnm{John} \sur{Shawe-Taylor}}

\author[5]{\fnm{Alessandro} \sur{Vespignani}}

\affil*[1]{\orgname{University of Pisa}, \country{Italy}}

\affil[2]{\orgname{Consiglio Nazionale delle Ricerche (CNR)}, \orgaddress{\city{Pisa}, \country{Italy}}}

\affil[3]{\orgname{Scuola Normale Superiore}, \orgaddress{\city{Pisa}, \country{Italy}}}

\affil[4]{\orgname{Sciences Po}, \orgaddress{\city{Paris}, \country{France}}}

\affil[5]{\orgname{Northeastern University}, \orgaddress{\city{Boston}, \country{USA}}}

\affil[6]{\orgname{Umeå University},
\orgaddress{\country{Sweden}}}

\affil[7]{\orgname{Central European University (CEU), \orgaddress{\city{Vienna}}, \country{Austria}}}

\affil[8]{\orgname{Victoria University Wellington, \orgaddress{\country{New Zealand}}}}

\affil[9]{\orgname{University of Athens}, \orgaddress{\country{Greece}}}

\affil[10]{\orgname{DFKI and University of Kaiserslautern}, \orgaddress{Germany}}

\affil[11]{\orgname{Massachusetts Institute of Technology (MIT)}, \orgaddress{\city{Boston}, \country{USA}}}

\affil[12]{\orgname{University College London}, \orgaddress{UK}}

\abstract{Human-AI coevolution, defined as a process in which humans and AI algorithms continuously influence each other, increasingly characterises our society, but is understudied in artificial intelligence and complexity science literature. Recommender systems and assistants play a prominent role in human-AI coevolution, as they permeate many facets of daily life and influence human choices on online platforms. The interaction between users and AI results in a potentially endless feedback loop, wherein users' choices generate data to train AI models, which, in turn, shape subsequent user preferences. This human-AI feedback loop has peculiar characteristics compared to traditional human-machine interaction and gives rise to complex and often ``unintended'' social outcomes.
This paper introduces \emph{Coevolution AI} as the cornerstone for a new field of study at the intersection between AI and complexity science focused on the theoretical, empirical, and mathematical investigation of the human-AI feedback loop. In doing so, we: \emph{(i)} outline the pros and cons of existing methodologies and highlight shortcomings and potential ways for capturing feedback loop mechanisms; \emph{(ii)} propose a reflection at the intersection between complexity science, AI and society; \emph{(iii)} provide real-world examples for different human-AI ecosystems; and \emph{(iv)} illustrate challenges to the creation of such a field of study, conceptualising them at increasing levels of abstraction, i.e., technical, epistemological, legal and socio-political.}




\maketitle

\newpage
\section{Introduction}
\label{sec:introduction}

\begin{description}
\item \small It is change, continuing change, inevitable change, that is the dominant factor in society today. No sensible decision can be made any longer without taking into account not only the world as it is, but the world as it will be (Isaac Asimov, \emph{Asimov on Science Fiction}, 1981) \cite{asimov1981asimov}. \\ 
\end{description}

The history of humankind is a history of coevolution: between humans and other species;  between humans and industrial machines; between humans and digital technologies; and, today, between  
humans and Artificial Intelligence (AI) \cite{russell2014coevolutionary, lee2020coevolution}.
Human-AI coevolution is a perpetual, iterative process wherein both humans and learning algorithms evolve in tandem, each influencing the evolution of the other over time. 
This generates complex effects on human-AI ecosystems and, therefore, on society \cite{geels2005co, lee2020coevolution, mokyr1995fourth, santosuosso2021coevolution}.
In this context, recommendation systems and assistants (in short, \emph{recommenders}) -- AI-based algorithms that suggest items or content based on users' preferences or specific requests \cite{ricci2015recommender, li2023recent, lu2015recommender} -- play a prominent role. 
Recommenders mediate, through online platforms, most actions in our daily lives by exerting instant influence over many specific choices.
Personalised suggestions on social media guide our content consumption and social connections, online retail recommenders propose products for consumption, navigation services suggest routes to reach our destinations, and generative AI creates content in response to users' wishes. 
We focus on recommenders because they wield unprecedented influence over human behaviour among AI applications. 
Unlike other AI tools, such as medical diagnostic support systems, robotic vision systems, or autonomous driving -- which assist in specific tasks or functions -- recommenders are ubiquitous in online platforms, shaping our decisions and interactions instantly and profoundly.
Therefore, studying the role of recommenders within human-AI ecosystems constitutes a vantage point to analyse the coevolution between humans and AI machines. 

\begin{figure}
\centering
\includegraphics[width=\textwidth]{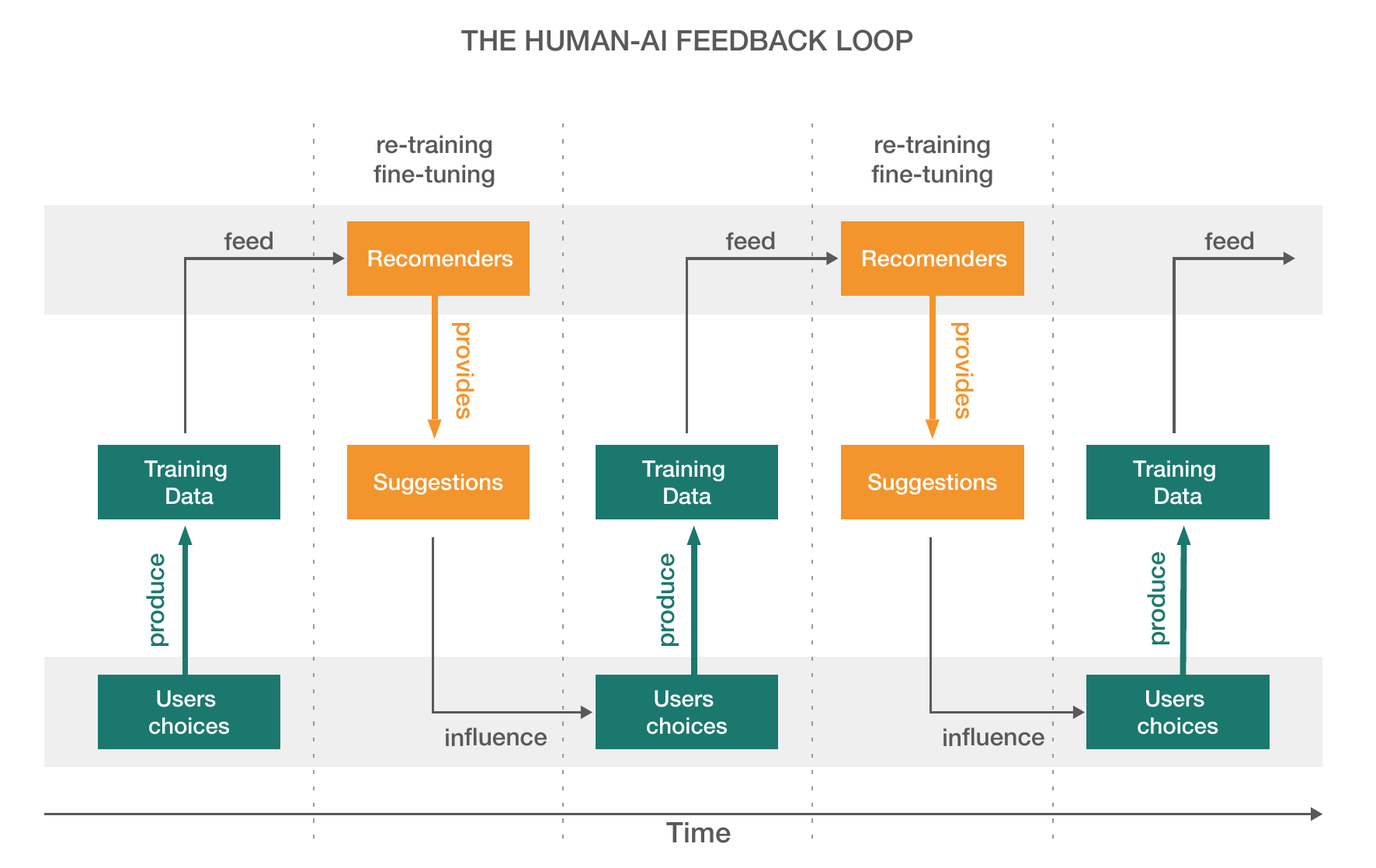}
\caption{Users' choices on online platforms generate data used to train recommenders. 
These recommenders then offer suggestions to users, influencing their choices, which in turn generate more data for re-training recommenders. 
This iterative process creates a potentially endless feedback loop.}
\label{fig:feedback_loop}
\end{figure}

Since recommenders are based on AI, and machine learning in particular, their interactions with users always give rise to a \emph{feedback loop} \cite{Wagner2021-hq,jiang2019degenerate,Sun2019-vl,Mansoury2020-zx}. 
One could describe the feedback loop as a process: users' choices determine the datasets on which recommenders are trained; the trained recommenders then exert an influence on users’ subsequent choices, which in turn affect the next round of training, initiating a potentially never-ending cycle (see Figure \ref{fig:feedback_loop}).  
The human-AI feedback loop may also lead to ``unintended'' social consequences \cite{Sirbu2019-bz,knott2021responsible,Isinkaye2015-xb, kulynich2020pots}.
Personalised recommendations on social media help users deal with information overload, but may artificially amplify echo chambers, filter bubbles, and processes of radicalisation \cite{Sirbu2019-bz,knott2021responsible,Del_Vicario2016-sd,Perra2019-vy, huszar2022algorithmic}. Profiling and targeted advertising may increase inequality and monopolies, perpetuating and accruing biases, discriminations, and the “tragedy of the commons” \cite{Pedreschi2008-cm,Kleinberg2020-jp,Mehrabi2021-qk, hosanagar2014will}. 
Navigation services suggest directions that make sense from an individual perspective, but may create chaos if too many drivers are sent to the same roads \cite{macfarlane2019navigation,Siuhi2016-gb,foderar2017navigation,cornacchia2022routing,Lima2016-hg,Colak2016-uw, cornacchia2023one, cornacchia2023navigation, kulynich2020pots}.  

While coevolution between humans and technology is not new, the human-AI feedback loop imbues it with forms that are unprecedented. 
Throughout history, technology and society have constantly coevolved, e.g., as testified by the advent of the press, the radio and the TV \cite{russell2014coevolutionary, lee2020coevolution}. 
Moreover, recommendations have always influenced human choices,  whether it was the role of a librarian picking up a book, a friend's advice on a music album to listen to, or the influence of TV advertisements on viewers. 
However, at least five fundamental aspects have been magnified by AI-based recommenders:
pervasiveness, persuasiveness, traceability, speed, and complexity.

Recommenders are pervasive in all online platforms, from social media to online retail and mapping services.
This integration, fueled by sophisticated algorithms, abundant data, and widespread user adoption, has established recommenders as customary elements of online interactions.
The large availability of data portraying individual choices empowers recommenders to deliver highly personalised suggestions.
This increases accuracy in capturing users' preferences, making recommenders highly persuasive.
Unlike previous technology, human-AI ecosystems leave an indelible trace of recommenders' suggestions and related human choices (what is commonly called big data). 
As a consequence, AI-based recommenders possess a comprehensive outlook of individual choices and an unprecedented capacity to shape human behaviour at scale.
Within this context, human-machine coevolution is faster than ever before, given that AI can be re-trained with little or no human oversight and provide suggestions at an unparalleled speed.
Human-AI ecosystems also foster an extraordinary volume of interactions between the huge spaces of users and products, escalating system complexity. 

The design of recommenders and the analysis of their impact are typically studied by two disciplines that gained prominence over the last two decades: artificial intelligence and complexity science.
However, working in isolation, these disciplines cannot address the challenges of understanding human-AI coevolution. 

AI has achieved human-like performance in many challenging tasks  \cite{Sejnowski2018-cr,LeCun2015-tu,Silver2016-tu,Titano2018-ox,Krizhevsky2017-ej,Wu2016-xa} and is becoming increasingly explainable and human-centric \cite{Guidotti2019-pt,Burkart2021-ce,Hofman2021-hl,Abdul2018-xo, Lukowicz2019-xp,Horvatic2021-qy,Lepri2021-od}.
However, it is still embedded in methodological individualism \cite{Dignum2022-ou}, where machines are studied as solitary agents and not from a coevolution perspective. 
We know little about the impact of the feedback loop on human-AI ecosystems; therefore, we need to disentangle whether recommenders amplify undesired collective outcomes or mitigate them.  
 
Complexity science has shown that networks of social interactions are heterogeneous, resulting in deeply connected hubs and modular structures 
\cite{Watts1998-zv,Barabasi1999-rn,Kleinberg2000-qq,Albert2002-mi,Onnela2007-xs,newman2018networks,barabasi2016network}. 
This structural heterogeneity interacts with social effects like inequalities and segregation, impacting on network processes such as the spreading of epidemics, information and opinions \cite{Sirbu2019-bz,Pastor-Satorras2015-dh,Colizza2007-qs}; the success of products, ideas, and people \cite{Wang2019-ph,Fraiberger2018-wq,Pappalardo2018-vu}; and urban dynamics \cite{brockmann2006scaling,gonzalez2008understanding,pappalardo2015returners,alessandretti2020scales,bohm2022gross,luca2021survey,barbosa2018human, simini2021deep,Hanna2017-ve, pappalardo2023future}.  
The human-AI feedback loop affects these network processes differently than before the advent of AI. 
A coevolutionary approach may help unveil the laws governing the complex interplay between humans and recommenders~\cite{Yang2020_HCI, Thurner2018Complexity, Eysinck2021AI_humans}, going beyond existing approaches~\cite{Lawless2023Editorial,Piao2023_H-AI_dyn,Contucci2022H-AI_Ising}.  
We know little about the parameters that govern the network dynamics of human-AI coevolution and, therefore, how to make predictions for such complex systems. 

This paper introduces Coevolution AI as the cornerstone for a new field of study at the intersection between AI and complexity science, and focuses on the theoretical, empirical, and mathematical investigation of the feedback loop.
Following in the footsteps of the literary imagination of Isaac Asimov, ``no sensible decision can be made any longer without taking into account not only the world as it is, but the world as it will be'' \cite{asimov1981asimov}. 
A detailed understanding of human-AI coevolution should be achieved by developing new methodologies and analytical frameworks.
Developing Coevolution AI is key for understanding how feedback loop mechanisms might impact societal dynamics, as the interaction between humans and recommenders might amplify or mitigate social phenomena. 
This is salient because we are immersed in a political economy that mainly privileges individual utility over collective goods, and where the means of production and recommendation are concentrated in a few hands. 

\begin{figure}
    \centering
    \includegraphics[width=\columnwidth]{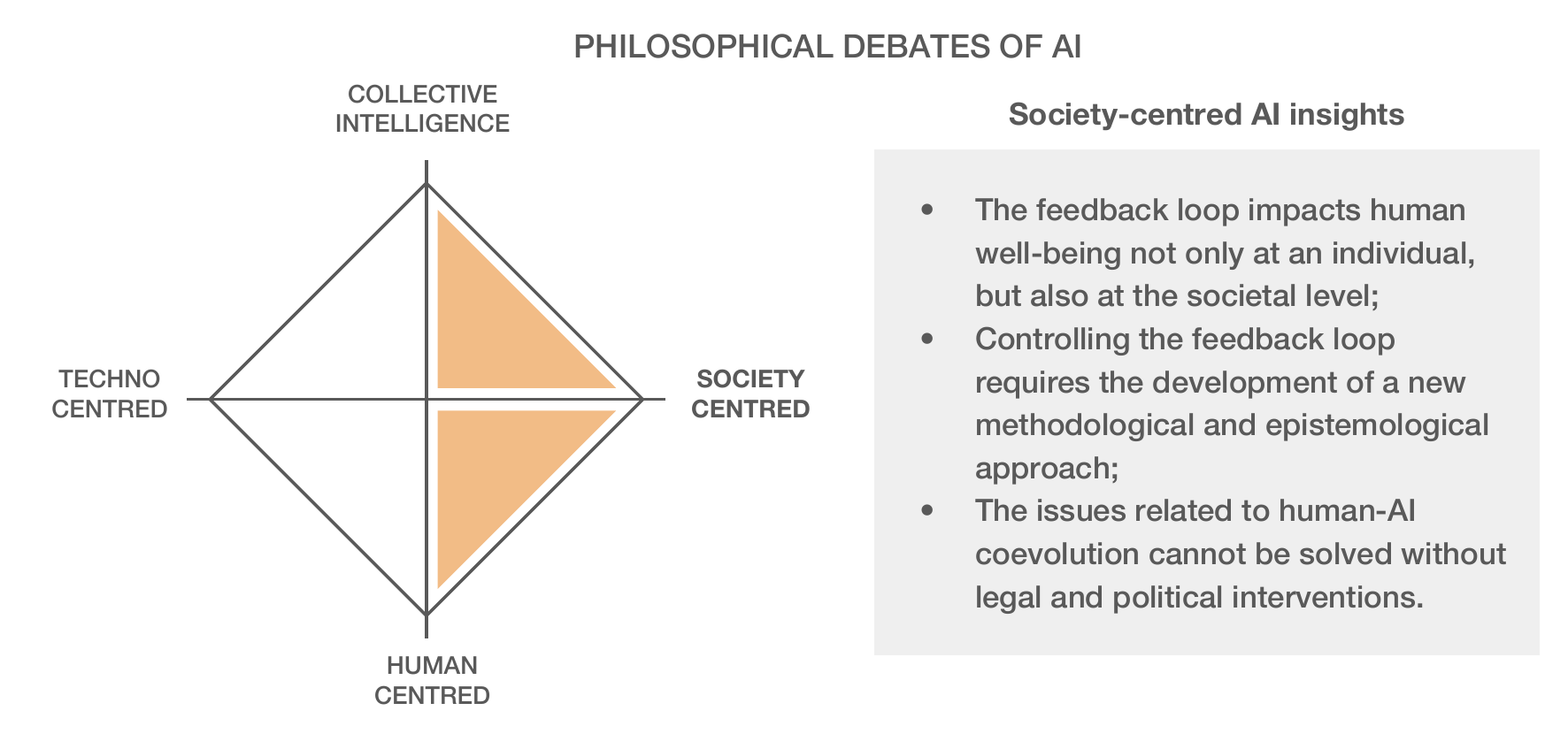}
    \caption{
    We enrich current philosophical perspectives about AI,  i.e., technology-centred AI, human-centred AI, and collective intelligence, with a new dimension that can be defined as society-centred AI.
    Society-centred AI brings three additional elements to the debate.
    }
    \label{fig:figure2}
\end{figure}

Coevolution AI sits at the intersection of two prominent AI debates. First, it expands the debate on ``machine behaviour'', i.e., the study of behaviour exhibited by intelligent machines \cite{rahwan2019machine}, illustrating that the feedback loop is the key process shaping human-AI coevolution. 
Second, it enriches current philosophical perspectives about AI, i.e., technology-centred AI, human-centred AI, and collective intelligence \cite{peeters2021hybrid}, with a new dimension that can be defined as society-centred AI. 
This perspective departs from technology-centred AI because it considers that the impacts of the feedback loop cannot be managed solely with additional AI technology. Moreover, it embraces seminal insights from human-centred AI and collective intelligence. 
On the one hand, from a human-centred AI perspective, it hypothesises that the feedback loop might curtail human well-being. 
On the other, from a collective intelligence perspective, it theorises how human-recommender interactions can drive coevolution towards desirable outcomes.
Society-centred AI brings three additional elements to the debate (see Figure \ref{fig:figure2}): \emph{(i)} the feedback loop impacts human well-being not only at an individual, but also at the societal level; \emph{(ii)} managing the feedback loop requires the development of new methodological and epistemological approaches; and \emph{(iii)} the issues related to human-AI coevolution cannot be solved without legal and political interventions. 

The remainder of the paper proceeds as follows. 
We review the methodologies employed in the literature to examine the interaction between humans and recommenders (Section \ref{sec:methods}). 
We then illustrate the outcomes of this interaction in four key ecosystems, i.e., social media, online retail, urban mapping, and generative AI ecosystems (Section \ref{sec:outcomes}) and speculate on the social impact of human-AI coevolution (in Section \ref{sec:social_impact}).
Finally, we outline open challenges in Coevolution AI, providing new avenues for future research (Section \ref{sec:conclusion}).

\section{Methods for human-AI coevolution}
\label{sec:methods}

This section provides a critical perspective on existing studies about the interaction between humans and recommenders and illustrates that only a few of them investigate the role of the human-AI feedback loop. 
In what follows, we analyse the pros and cons of each methodology employed and highlight shortcomings and potential ways for capturing feedback loop mechanisms.

\subsubsection*{Empirical vs Simulation studies} Existing studies mostly employ either an empirical or a simulation approach.
Empirical studies are based on data generated by users' behaviour on real platforms, which emerge from dynamic interactions between humans and recommenders.  
These studies provide evidence about specific phenomena and an empirical basis for their understanding. When based on large samples or diverse datasets, empirical studies also enable researchers to make generalisations about broader populations. 
Nonetheless, the possibility to draw general conclusions is limited by the specific time frames and conditions under which these studies are conducted. 
In addition, as data are primarily owned by big tech platforms and rarely made public, these studies are hardly reproducible. 

Simulation studies are based on data generated by mechanistic, AI-, or digital-twin-based models. 
Simulation studies provide a cost-effective alternative to empirical studies, especially when dealing with large-scale ecosystems or when data is not readily available. 
They can be reproduced with the same initial conditions, allowing for the verification and validation of results. 
Scholars can manipulate parameters to observe the effects on the human-AI ecosystem, helping the research community understand the intricate relationships between variables. 
However, as they are based on heavy assumptions, simulations do not necessarily reflect real-world dynamics and are, therefore, limited in unveiling unexpected or unintended outcomes. 
The settings of specific parameters may close the door to unforeseen results.  

\subsubsection*{Controlled vs observational studies}
Both empirical and simulation approaches can be either observational or controlled. 
Controlled studies, called experiments in social sciences \cite{deaton2018understanding, hariton2018randomised}, include quasi-experiments, randomised controlled trials and A/B tests. 
These studies split the sample into a control group and one or more experimental groups, each subjected to different recommendations \cite{deaton2018understanding}. 
The influence of recommendations is measured by comparing the outcomes obtained in the two (or more) groups. 
Controlled studies allow researchers to control for various factors and conditions, facilitating the isolation of the effect produced by a specific intervening variable. 
Their main advantage is to establish causal relationships and attribute observed effects to the recommendation. 
Moreover, sample randomization reduces selection biases, ensuring that participants in both groups have an equal chance of receiving the recommendation. 
However, controlled studies also have important shortcomings: the inclusion and exclusion criteria of the controlled settings might limit the generalisability of findings; and there is limited flexibility in adapting to changes intercurring during the experiments. 
Moreover, they are hard to design because they require direct access to platforms' users and recommenders \cite{knott2022transparency}.

Some examples of controlled studies can help clarify the characteristics of these methodologies.
\citet{huszar2022algorithmic} document an experiment that investigates the effect of Twitter personalised timeline on the diffusion of political content from different parties. 
The platform selected a group of users that were exposed to tweets in reverse-chronological order from accounts they followed, while a treatment group of users was exposed to the AI-based personalised recommender. 
The personalised recommendations resulted in a substantial amplification of political messages, with mainstream right-wing parties benefiting more from algorithmic personalisation than their left-wing counterparts. 
\citet{cornacchia2022routing} used a traffic simulator to estimate the impact of real-time navigation services on the urban environment. 
The study found that navigation services may increase travel time and CO2 emissions when the adoption rate crosses a threshold.

Observational studies assume a single recommendation principle without any control \cite{pera2023measuring}. 
Examples include the analysis of Facebook users' behaviour \cite{bakshy2015exposure}, Google Maps' suggestions to drivers \cite{Arora2021-xc}, data gathered from browser loggers \cite{Flaxman2016-xa, Hosseinmardi2020-wb}, platforms’ APIs \cite{ledwich2019algorithmic, bakshy2015exposure}, bots that simulate human behaviour \cite{Whittaker2021-sb, Ribeiro2020-jl, Papadamou2021-ns}, and experiments that ask volunteers to behave in specific ways \cite{Allcott2019-lt, Asimovic2021-ap, Levy_undated-lj, cho2020search}. 
A strength of observational studies (either based on empirical data or simulations) is that, when data is big and representative enough, they allow researchers to make generalisations about a broader population, enhancing the external validity of findings and highlighting potential biases towards different population segments. 
A significant limitation, however, is that establishing a causal relationship is challenging; additional evidence is required to support causal claims. 
Moreover, findings of observational studies may suffer from selection biases, measurement errors, or the presence of confounding variables; this may compromise their accuracy and reliability. 

To clarify the characteristics of observational studies, we rely on two examples. 
\citet{cho2020search} analysed factors leading to political polarisation on YouTube, finding that a recommender can contribute to this polarisation, but mainly through the channel of a user's preference. 
\citet{fleder2011recommender} investigated the influence of recommenders on users' purchases on a music streaming platform. 
They found that users exposed to recommendations purchase more items and are more similar to one another in the variety of what they buy.

\subsubsection*{Modelling the feedback loop}
There is a handful of studies that move beyond the static investigation of the influence of recommenders on users' behaviour, analysing feedback loop mechanisms theoretically and/or empirically \cite{ensign2018runaway, jiang2019degenerate, Mansoury2020-zx, Sun2019-vl, nguyen2014exploring}.
Some of these works introduce mathematical models to provide insights on recommenders' influence based on their parameters.
\citet{jiang2019degenerate} theoretically investigated whether feedback loop mechanisms lead to a degeneration of users' interest: 
an oracle recommender (with perfect accuracy) induces a quick degeneration, while injecting randomness in users' choices and enlarging the pool of items slows this process down. 
Therefore, this study provides insights and potential remedies against the degeneration of the feedback loop.

There is also research that combines a theoretical and an empirical appraisal of the feedback loop.
\citet{ensign2018runaway} investigated the impact of a predictive policing recommender, based on historical crime data, on distributing police resources among urban districts. 
They simulated a scenario where the following feedback loop takes place: officers are deployed daily to districts with the highest predicted crime rate, crimes discovered by these officers are reported, and then data about these reported crimes are fed back into the recommender.
The process continues iteratively.
Given this feedback loop, the recommender repeatedly redirects police attention to districts where more crimes are reported.
As more crimes are likely to be discovered in these districts due to the increased presence of officers, in the long run, the simulation leads to a distribution of crimes that is unrealistic if compared with observed historical crime data. 
The authors proposed a correction mechanism where the likelihood of police deployment to a district decreases as discovered crime data are incorporated into the recommender. 
Simple urn models were employed to model the feedback loop and the correction mechanism.
One may interpret this evidence as suggesting that small changes in the recommender can make a real difference in the human-AI ecosystem. 

Although these seminal studies provide intriguing insights into feedback loop mechanisms, there are important avenues for further improvement in the analysis of human-AI coevolution.
At the empirical level, data employed in these studies allow drawing only an incomplete picture of the interaction between humans and recommenders. 
Typically, they only describe users' choices at time $t_i$, without considering which recommendations are provided to influence users' choices at $t_{i - 1}$. 
Moreover, we have no information about how often platforms re-train recommenders on users' choices to update the knowledge of their preferences.
For these reasons, existing studies employ available data only to validate theoretical approaches that model feedback loop mechanisms. 
To overcome this limitation, we need empirical studies based on longitudinal data describing, at each iteration of the feedback loop: \emph{(i)} the recommendations provided to users; \emph{(ii)} the reaction of users to these recommendations; and \emph{(iii)} the process of re-training based on users' choices influenced by previous recommendations.
Studies of this kind would permit looking at causation bi-directionally and not only unidirectionally.

\section{Outcomes of human-AI coevolution}
\label{sec:outcomes}
An outcome can be defined as the effect of the influence recommenders and users exert on human-AI ecosystems, and can be measured at different levels, i.e., individual, item, model, and systemic. 
Individual outcomes refer to the influence of recommenders on users, e.g., sellers and buyers in the online retail ecosystem, and drivers and passengers in the urban mapping ecosystem. 
Item outcomes refer to the influence of recommenders and users' choices on the characteristics of specific objects; examples include posts on social media, products on online retail platforms, rides in urban services, and generated content on the generative AI ecosystem. 
Model outcomes pertain to the influence of users' choices on the characteristics of the recommender. 
This encompasses whether the recommender alters its behaviour and the nature of its recommendations in response to users' choices.
Systemic outcomes refer to the collective impact of the interaction between humans and recommenders.

Reviewing the literature, we detect several systemic outcomes, such as polarization, echo chambers, inequality, concentration, and segregation. Polarization is a sharp separation of users or items into groups based on some attributes (opinions or beliefs) \cite{Sirbu2019-bz, valensise2023drivers, pansanella2022mean, peralta2021effect, peralta2021opinion, pansanella2023mass, liu2023algorithmic, chitra2020analyzing, haroon2023auditing, bouchaud2023crowdsourced, cho2020search, guess2023social, alvim2023formal, rahaman2021model, yang2023bubbles, pansanella2022modeling, Perra2019-vy, Ribeiro2020-jl}. 
Echo chambers are environments in which opinions or item choices within a group are confirmed and reinforced \cite{chitra2020analyzing, Hosseinmardi2020-wb, Perra2019-vy, cinus2022effect}. 
Inequality indicates an uneven distribution of resources among members of a group, while concentration is a close gathering of users or items \cite{ledwich2019algorithmic, boeker2022empirical, bartley2021auditing, fabbri2022exposure, kirdemir2021assessing, ng2023exploring}. 
Concentration is typically identified as congestion in the urban context \cite{Hanna2017-ve, erhardt2019transportation, cornacchia2023one, cornacchia2022routing, cornacchia2023navigation, Arora2021-xc, falek2022reroute, santi2014quantifying, fagnant2018dynamic, jalali2017emission, kulynich2020pots}. 
Segregation is a situation in which groups of users are set apart from each other \cite{zhang2021frontiers, koh2019offline}. Some outcomes emerge at the individual or model level only.
Examples are filter bubbles (individual level), which is a conformation of contents with a user's own beliefs \cite{srba2023auditing, tomlein2021audit, Abdul2018-xo, grossetti2019community}, and model collapse (model level), i.e., a deterioration of the recommender's performance as it continues to coevolve with users \cite{shumailov2023curse,  guo2023curious, briesch2023large, dohmatob2024tale, alemohammad2023self, martinez2023towards, dohmatob2024model, bohacek2023nepotistically, hataya2023will}. 
Finally, outcomes emerging at all levels are changes in volume (i.e., a quantity measuring some users' or items' attribute) and changes in diversity, which can be defined as the variety of items and users' behaviours \cite{pathak2010empirical, chen2004impact, donnelly2021long, chen2022more, yi2022recommendation, matt2013differences, alves2023digitally, yi2022recommendation, noordeh2020echo, aridor2020deconstructing}. 

All these outcomes emerge, often unintendedly, as the result of the human-AI coevolution. 
A notable example is the emergence of concentration, conformism and diversity loss. 
In what follows, we discuss other examples gathered from four largely studied human-AI ecosystems, i.e., social media, online retail, urban mapping, and generative AI ecosystems.

\textbf{Social media.} Recommenders have achieved considerable success in the realm of social media, where they are deployed to suggest new posts and users to follow. 
This coevolution between users and recommenders gives rise to two interconnected feedback loops. First, previous interactions between users and posts shape actual recommendations; these recommendations influence subsequent interactions between users and future posts. 
Second, the accounts a user follows shape actual user recommendations; these recommendations influence subsequent interactions between users and their followers.
Given that social media disseminate opinions, this feedback loop can lead to various outcomes, at individual (filter bubble, homophily) and systemic levels (polarisation, fragmentation, echo chamber). 
While recommenders assist users in accessing content and connecting with like-minded individuals, the underlying algorithms may confine them within a filter bubble. 
This confinement can contribute to significant polarisation of opinions and users, fostering the potential for radicalization of ideas. 
For example, as previously discussed, personalised recommendations on Twitter overexpose users to certain political content \cite{huszar2022algorithmic}.

\textbf{Online retail.} Recommenders also play a pivotal role in the success of e-commerce and streaming giants like Amazon, eBay, and Netflix. 
The coevolution of these recommenders with consumers may give rise to intricate feedback loop mechanisms. 
The recommended items (e.g., consumer goods, songs, movies) depend on previous purchases, which in turn were influenced by previous recommendations. 
A crucial distinction in this human-AI ecosystem lies between collaborative filtering and personalised recommenders. 
Collaborative filtering operates on the principle of ``who-buys-this-also-buys-that'', relying on collective user behaviour; while personalised recommendations tailor suggestions to individual user taste. 
For example, collaborative filtering may increase sales volume and individual consumption variety while, at the same time, may decrease overall aggregate consumption diversity, amplifying the success of popular products \cite{fleder2009blockbuster,Lee2019-xr}. On the one hand, recommenders help users better navigate the large space of product choice, reducing effort and choice overload, allowing a rapid allocation of needed goods and boosting platform revenues. 
On the other, at the aggregate level, they might reduce the variety of purchased products (i.e., creating filter bubbles around the user) and increase concentration, favouring certain brands and reducing competition. 

\textbf{Urban mapping.} Navigation services recommend a route to a destination, considering changing traffic conditions and assisting users in exploring unfamiliar areas. 
Therefore, users with the same origin and destination receive similar recommendations. 
The impact of navigation services on the city is unclear: being designed to optimise individual travel times, they may also cause congestion and lead to longer travel times and higher CO2 emissions in the environment \cite{macfarlane2019navigation, Siuhi2016-gb, cornacchia2022routing, cornacchia2023one, cornacchia2023navigation, kulynich2020pots}. 
For example, in 2017, Google Maps, Waze and Apple Maps re-routed drivers from congested highways to the narrow and hilly streets of Leonia (a small town in New Jersey), creating such congestion that people could not get out of their driveways \cite{foderar2017navigation, macfarlane2019navigation, kulynich2020pots}. 
These issues are exacerbated by the coevolution between drivers and algorithmic updates, generating a feedback loop: travel times, which shape actual recommendations, also depend on drivers’ route choices that were influenced by previous recommendations. Drivers’ behaviour may change travel times in return, shaping subsequent recommendations. 
In this context, if too many drivers choose the same ``eco-friendly'' route, this route will cease to be eco-friendly. 

\textbf{Generative AI.} Recently emerging large language models (LLMs), deployed by big tech companies (e.g., Amazon, Baidu, Google, Meta, Microsoft, OpenAI) and/or open source projects (e.g., Bloom, Cerebras-GPT, Dolly, Falcon, Mistral, Zephyr), are rapidly permeating various domains, e.g., education, politics, work. 
It has been observed that LLMs may potentially compress diversity, thus standardising the use of language in generated text \cite{shumailov2023curse,  guo2023curious, briesch2023large, dohmatob2024tale, alemohammad2023self, martinez2023towards, dohmatob2024model, bohacek2023nepotistically, hataya2023will}. 
Recent studies show that when LLM-generated content is used to fine-tune LLMs themselves, a ``regression to the mean'' can take place with a loss of linguistic diversity in the form and substance of the generated text. 
This process of autophagy will be increasingly widespread as content on the web, which is the fuel that trains and fine-tunes LLMs will be more likely to be generated by machines rather than humans.  
At an analytical level, we need to understand this coevolution to develop recommenders that balance the feedback loop's potential positive (standardisation of language) and negative impacts (compression of linguistic diversity). 

\section{Social impact of human-AI coevolution}
\label{sec:social_impact}
With our concept of Coevolution AI, we do not simply propose a reflection on the interconnection between complex systems and AI, but also on their relationship with society. Some of the outcomes discussed above have a long history in social science. Notable examples are polarization (in political science \cite{hetherington2009putting, abramowitz2008polarization}), inequality (in economics \cite{atkinson1970measurement}, in sociology \cite{beck2007beyond}), and segregation (in economics \cite{schelling1969models}, in urban studies \cite{jacobs1961death}).

The fact that recommenders have substantial and often ``unintended'' consequences on these social phenomena is central to Coevolution AI, as they might amplify trends already in motion in society. 
Studies in different social science fields (e.g., political science, sociology, political economy, economics and psychology) have shown that the increase of individualism \cite{santos2017global} and inequalities \cite{milanovic2016global, piketty2014capital} and the retrenchment of public and social policies \cite{ferragina2022welfare, ferragina2019political, ferragina2024two, ferragina2022rising, ferragina2021selective, ferragina2022labour} are tangible realities across several countries. 
In this respect, Coevolution AI as a field of study should also integrate insights from social sciences to evaluate potential societal consequences of feedback loop mechanisms. 
For reasons of brevity, we will put the spotlight on three examples of such an intersection, but our considerations extend well beyond: \emph{(i)} individual utility vs common goods; \emph{(ii)} ownership of ``the means of recommendation''; and \emph{(iii)} phenomena of inequality and concentration. 

Research on recommenders is mostly designed for the maximisation of individual utility and profit for companies (e.g., \cite{zhang2021frontiers, garcia2020short} for the urban mapping ecosystem, \cite{dias2008value, chen2004impact, hosanagar2014will, yi2022recommendation, lee2014impact} for the online retail ecosystem), and this generates a lack of consideration for common goods. 
A good starting point to understand the classical limitations of this approach is the debate on rational choice theory and methodological individualism \cite{oppenheimer2008rational, kjosavik2003methodological}. 
Rational choice theory is grounded in the idea that individual agents act in society by following rational reasoning, and therefore maximising their utility. 
Methodological individualism is a sort of corollary and explains societal outcomes as the sum of the relational behaviours of all agents within society. 
Recommenders designed on the basis of methodological individualism might not take into account collective utility. 

History offers plenty of examples of societal damage generated by the prevalence of individualistic over collective behaviours. 
One could go back to Thomas More and ``the drama of enclosures'' to conceptualise the potential paradoxes that the coevolution between humans and recommenders might generate. 
The enclosures were a vast movement, which privatised land that had been used collectively. 
Similarly, employing individual utility as the main driver to set up recommenders can reduce the potential collective utility provided by the development of AI, and therefore, reinforce polarisation mechanisms already existent in society.  

Human-AI coevolution is connected to the functioning of capitalism: the dramatic technological developments related to recommenders and AI are taking place in a period dominated by rational choice theory, methodological individualism and ultimately neoliberal economics \cite{harvey2007brief}. 
Our reflection extends Kean Birch’s notion of ``automated neoliberalism'' \cite{birch2020automated}, which posits that digital platforms shape markets, personal data accumulation transforms individual lives, and algorithms have the potential to automate social relations. 

The potential for improving collective utility might be reduced because the coevolution between AI and society takes place in a context where there is no balance between those who own recommenders and those who use them uncritically. The old question about who owns the means of production strikingly applies today and takes a new form in an environment of Coevolution AI. 
Readapting Marx’s reasoning \cite{marx2011capital}, who owns the ``means of recommendations''? How do these recommendations affect subsequent interactions between humans and AI? How could this reinforce outcomes such as inequalities and polarisation? 

It is not the scope of this perspective article to delineate a political economy of Coevolution AI \cite{trajtenberg2018ai}, intended as a systematic analysis of the interconnection between asset ownership and the influence of recommenders. 
However, this aspect is a fundamental contextual element to discuss, at least theoretically, the potential unintended consequences of the coevolution between humans and recommenders. 
If recommenders have a powerful influence in shaping individual choices, and these individual choices shape collective outcomes, we have to reflect on how these choices are taken. 
Moreover, the impact of coevolution can further strengthen outcomes and increasingly harm collective utility. 
The opposite reasoning is also valid, and a different socio-economic context, where recommenders are geared towards shared collective objectives, can foster positive societal outcomes through coevolution. 
This point underlines that mere techno-solutionism \cite{morozov2013save}, the idea that technology is the answer to any challenge we face, is a risky intellectual posture, and we need more studies on human-AI coevolution to steer positive outcomes.
A review of recent research shows that AI tendentially increases the social divide, especially for historically marginalised groups (e.g., research about the US points to racial and gender effects) \cite{sartori2022sociotechnical}. 
These patterns are even stronger in low/middle income countries \cite{hagerty2019global}. 
One could reasonably hypothesise that the uneven impacts of AI are underestimated because we miss research measuring coevolution and feedback loop mechanisms.   

Human-AI coevolution might, therefore, foster inequality and concentration across different ecosystems.  
Within the social media ecosystem, recommenders may intensify exposure inequalities at the individual level, exacerbating the rich-get-richer effect regardless of user attributes or network characteristics \cite{fabbri2022exposure}. 
Concerning the online retail ecosystem, purchase-based collaborative filtering might increase the number of items purchased and the variety of products considered. 
However, this might also push users to buy the same products, contributing to concentration at the systemic level \cite{lee2014impact}. 
In the urban mapping ecosystem, ride-hailing recommenders (like Uber and Lyft) may mislead low ridership in poor and black neighbourhoods as a reflection of low users' demand, reinforcing in this way existing racial and socio-economic inequalities \cite{ge2020racial, Yan2020-er}.
The autophagy process in a generative AI ecosystem may lead to diversity loss in the generated content \cite{shumailov2023curse}. 

\section{Open Challenges of Coevolution AI}
\label{sec:conclusion}
\begin{figure}
    \centering
    \includegraphics[width=0.8\columnwidth]{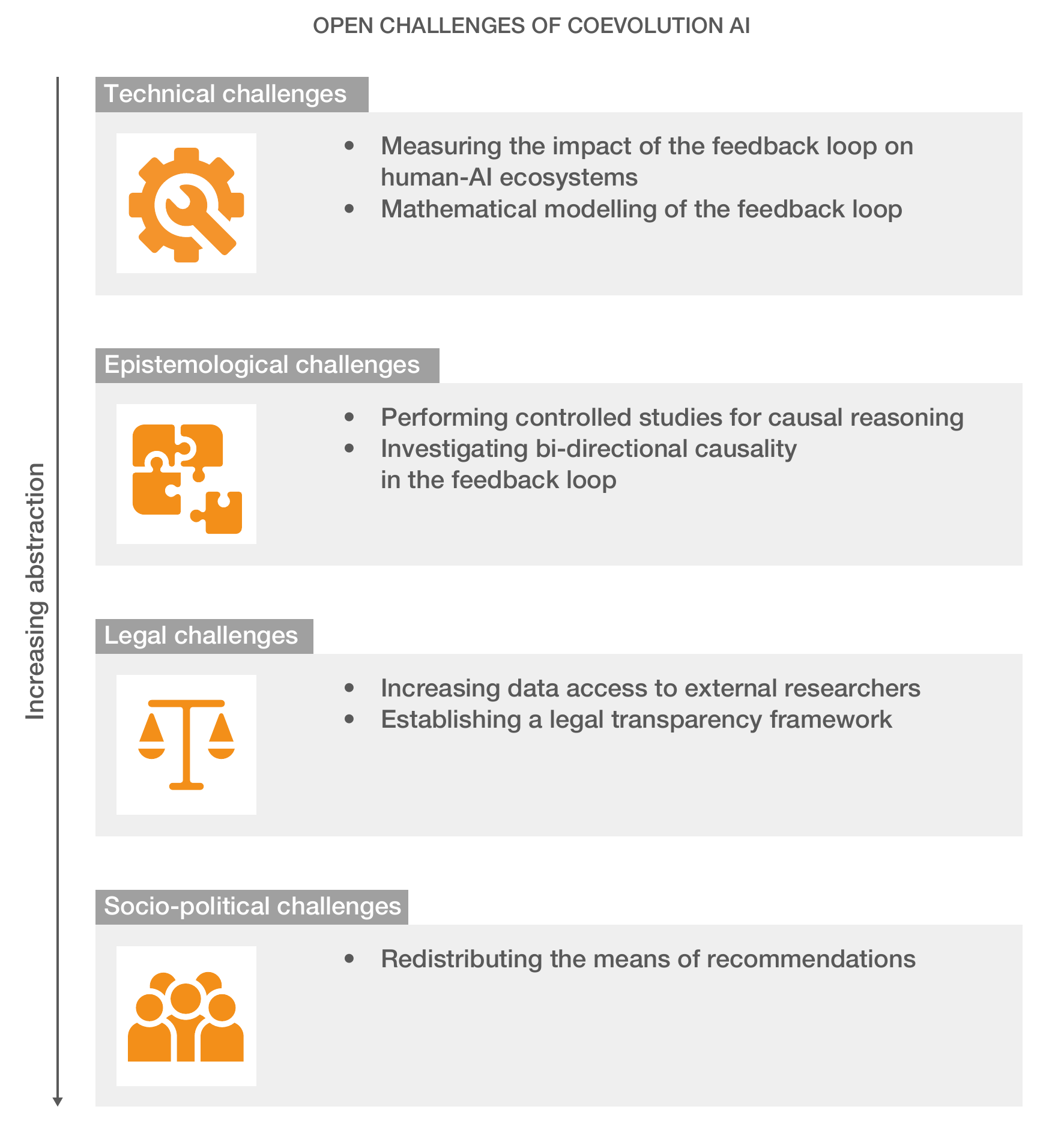}
    \caption{Coevolution AI presents important challenges for the future, that can be conceptualised at increasing levels of abstraction: technical, epistemological, legal and socio-political.}
    \label{fig:figure3}
\end{figure}

Coevolution AI presents important challenges for the future that can be conceptualised at increasing levels of abstraction, i.e., technical, epistemological, legal and socio-political (see Figure \ref{fig:figure3}).  

From a technical perspective, we need a method to continually measure the impact of the feedback loop on the behaviour of humans and recommenders. 
Such a method could be developed by tracking step-wise how the measured outcomes change every time the recommender is re-trained. 
For example, one can measure the variety of products purchased by users at time $t_0$, assessing the extent to which feedback loop mechanisms might amplify or reduce this variety with successive steps, e.g., $t_1, t_2, \dots, t_n$.
How many successive feedback loop iterations might be required to alter this variety of purchases?
In a similar fashion, we could track how the recommender changes after subsequent re-training rounds: how long it takes a generative AI model to collapse and lose linguistic diversity? The conformism/diversity balance is yet another fundamental aspect related to the feedback loop that deserves more attention.  
At a more general level, we also need a conscious effort in mathematical modelling to capture feedback loop mechanisms and their impact on human-AI ecosystems.

From an epistemological perspective, understanding the causal interplay between humans and recommenders is key; however, much of the existing literature overlooks coevolution. 
Some steps could be undertaken to mitigate this issue, such as the development of controlled studies accounting for feedback loop mechanisms.
Moreover, we must move beyond a unidirectional view of causality and explore it bi-directionally: humans and recommenders exert continuous influence on each other, necessitating a holistic study of their coevolutionary dynamics.

Beyond methodological and epistemological challenges, other barriers can prevent our capacity to study the emerging phenomenon of human-AI coevolution, e.g., the limited access to data for researchers that are external to platforms and the lack of transparency on how recommenders are designed and employed within different platforms \cite{knott2021responsible, knott2022transparency}. 
This severely impairs the reproducibility and replicability of potential Coevolution AI studies.  
Initiatives like EU's Digital Services Act might mitigate this barrier, but it remains unclear how vetted researchers will be allowed to access privately-owned platforms. 
Besides a new legal transparency framework, an intriguing way to overtake these barriers might involve the development of specialized APIs. 
APIs can allow external researchers to interact with platforms, and conduct empirical controlled experiments by changing recommenders' parameters or building experimental and control groups of users.
Governments should foster a culture of impact evaluation of feedback loop mechanisms among platforms. 
This might follow the model already in place to reduce and compensate for other forms of negative externalities, e.g., pollution and drug undesired effects.

Increasing transparency also requires dealing with other fundamental challenges at the socio-political level. 
The concentration of what we defined as ``the means of recommendations'' is crucial in this perspective. 
In a context where big tech companies enjoy a situation of oligopoly, recommenders are calibrated to generate high profits for the few. 
In this respect, a socially desirable development of Coevolution AI might be hindered by the lack of political intervention to redistribute the means of recommendation across markets with many smaller players and, more broadly, society. 
Such a configuration could be conducive to developing more transparent rules in data access and a fairer distribution of the means of recommendation. 
In the long term, small changes in the functioning of recommenders or humans' behaviour might lead to a significant impact on social outcomes, be they positive or negative. 
This could create a sort of ``butterfly effect'' of the feedback loop \cite{bradbury2016sound} that we have to study and understand.

The discussed challenges will persist even in the case of potential deep transformations in future online platforms, such as the emergence of decentralised platform architectures, user ownership of data and AI agents, and novel models for platform oversight and governance.
These challenges not only concern scholars and their capacity to do research, but also go well beyond and bear upon the social and political realm.
Only by accurately measuring and understanding the influence of recommenders on human behaviour we will be able to inform policymakers on how to make sensible decisions ``taking into account not only the world as it is, but the world as it will be'' \cite{asimov1981asimov}. 
This understanding will be key to designing adequate policies to avoid the potential negative externalities of an uncontrolled coevolution between humans and recommenders.
We aim at a future society-centric AI that is part of the solution to long-standing societal problems, instead of being part of the problem.

\bmhead{Acknowledgments}
This work has been partially supported by:
\begin{itemize}
    \item PNRR - M4C2 - Investimento 1.3, Partenariato Esteso PE00000013 - "FAIR - Future Artificial Intelligence Research" - Spoke 1 "Human-centered AI", funded by the European Commission under the NextGeneration EU programme;
    \item EU project H2020 HumaneAI-net G.A. 952026;
    \item EU project H2020 SoBigData++ G.A. 871042;
    \item ERC-2018-ADG G.A. 834756 “XAI: Science and technology for the eXplanation of AI decision making”;
    \item CHIST-ERA grant CHIST-ERA-19-XAI-010, by MUR (grant No. not yet available), FWF (grant No. I 5205), EPSRC (grant No. EP/V055712/1), NCN (grant No. 2020/02/Y/ST6/00064), ETAg (grant No. SLTAT21096), BNSF (grant No. K$\Pi$-06-AOO2/5);
    \item Luca Pappalardo has been supported by PNRR (Piano Nazionale di Ripresa e Resilienza) in the context of the research program 20224CZ5X4\_PE6\_PRIN 2022 “URBAI -- Urban Artificial Intelligence” (CUP B53D23012770006), funded by European Union -- Next Generation EU;
    \item Emanuele Ferragina has been partially supported by a public grant overseen by the French National Research Agency (ANR) as part of the ‘Investissements d’Avenir’ program LIEPP (ANR-11-LABX-0091, ANR-11-IDEX-0005-02) and the Université de Paris IdEx (ANR-18-IDEX-0001).
\end{itemize}
We thank Vincenzo Vivarini for the inspiration about the importance of the feedback loop in complex social systems. There is always a human agent beyond any tactical system. 
We also thank Daniele Fadda for the support on making the figures.
Finally, we thank all members of KDD-Lab for insightful discussions about Coevolution AI.

\bibliography{biblio}

\end{document}